\documentclass[letterpaper]{article} %
\usepackage{aaai2027}
\nocopyright %
\usepackage[hyphens]{url}  %
\usepackage{graphicx} %
\usepackage{xcolor}
\definecolor{diffdel}{RGB}{255,235,235}
\definecolor{diffadd}{RGB}{232,248,232}
\usepackage{natbib}  %
\usepackage{caption} %
\usepackage{algorithm}
\usepackage{algorithmic}

\usepackage{newfloat}
\usepackage{listings}
\DeclareCaptionStyle{ruled}{labelfont=normalfont,labelsep=colon,strut=off} %
\floatstyle{ruled}
\newfloat{listing}{tb}{lst}{}
\floatname{listing}{Listing}

\usepackage{booktabs}
\usepackage{array}
\usepackage{amssymb}
\usepackage{mdframed}
\newmdenv[linecolor=black!55,linewidth=0.6pt,backgroundcolor=black!4,
  innerleftmargin=6pt,innerrightmargin=6pt,innertopmargin=6pt,innerbottommargin=6pt,
  skipabove=8pt,skipbelow=8pt]{casebox}
\title{OPAL: Optimizing Proofs at Library Scale via Metric-Driven Coding Agents}
\author{
    Chengwu Liu\textsuperscript{\rm 1},
    Zijin Feng\textsuperscript{\rm 2},
    Ye Yuan\textsuperscript{\rm 1},
    Yifang Qin\textsuperscript{\rm 1},
    Botao Li\textsuperscript{\rm 1},
    Jiaxuan Xie\textsuperscript{\rm 1},
    Siqi Li\textsuperscript{\rm 1},\\
    Pu Suo\textsuperscript{\rm 3},
    Haoli Bai\textsuperscript{\rm 2},
    Yichun Yin\textsuperscript{\rm 2},
    Lifeng Shang\textsuperscript{\rm 2},
    Ming Zhang\textsuperscript{\rm 1}\corresponding
}
\affiliations{
    \textsuperscript{\rm 1}State Key Laboratory of Multimedia Information Processing, School of Computer Science,\\
    PKU-Anker LLM Lab, Beijing Key Laboratory of Software and Hardware Cooperative\\
    Artificial Intelligence Systems, Peking University\\
    \textsuperscript{\rm 2}Huawei Technologies Co., Ltd.\\
    \textsuperscript{\rm 3}Georgia Institute of Technology\\
    liuchengwu@pku.edu.cn, mzhang\_cs@pku.edu.cn
}

\begin{document}

\maketitle

\begin{abstract}
Driven by rapid progress in autoformalization and automated theorem proving, formal mathematics is scaling up: mathematicians, often with AI assistance, advance large formalization projects around Lean blueprints, which tightly link a \LaTeX{} proof plan to the Lean codebase. As mathlib and proof automation evolve, finished proofs accumulate optimization opportunities. Yet even in the mature mathlib library, 41\% of commits add no new mathematics but improve existing proofs (shortening, deduplication, faster compilation), and 62\% of these span multiple files, beyond the reach of theorem-level methods. We introduce OPAL, the first repository-level, metric-driven proof-optimization task, where an agent freely refactors a blueprint repository toward repository-level objectives. We instantiate OPAL with six metrics grounded in empirically mined human maintenance tendencies, including proof length, import closure, mathlib duplication, and elaboration cost. Optimizing these metrics freely, however, invites reward hacking, such as deleting a theorem to shorten the code, so evaluation needs a contract on what must stay intact. Our key observation is that blueprint structure itself supplies this contract: \texttt{\textbackslash lean\{\}} links pin down which theorems constitute a project's final targets (5--19\% of all declarations), whose statements must be preserved, while the remaining 81--95\% of auxiliary code may be freely refactored. We implement this contract as an executable semantic-preservation checker. Building on it, we release OPAL-Bench, comprising 162 task instances from 18 real-world Lean blueprint projects, together with a sandboxed agent harness. Experiments show that coding agents can deliver genuine optimization gains beyond the level of single human commits, while ablations show that with the contract absent from the agent's world nearly half of accepted improvements do not survive semantic verification.
\end{abstract}

\begin{links}
    \link{Code}{https://github.com/liuchengwucn/opal}
    \link{Data}{https://hf.co/datasets/liuchengwu/OPAL-Bench}
\end{links}

\section{Introduction}

Formal mathematics is moving past the bottleneck of expert manual formalization and beginning to scale. The Lean ecosystem centered on mathlib has accumulated over 2.4 million lines of proof code~\cite{lean4,mathlib}. A further notable development is the Lean blueprint workflow, in which mathematicians start from a \LaTeX{} proof plan, bind each lemma to a Lean declaration via \texttt{\textbackslash lean\{\}} links, and then, often with AI assistance, gradually fill in the formal proofs~\cite{leanblueprint}. Large projects such as FLT and Carleson are advancing rapidly in this fashion~\cite{flt,carleson}, and progress in autoformalization and automated theorem proving continues to accelerate this process~\cite{alphaproof,deepseekprover}.

Completing the proofs, however, is only the beginning of a formal library's life cycle. Our analysis of more than 30{,}000 mathlib commits shows that, in the stable era after the Lean~4 port, 41.4\% add no new mathematics but instead improve existing proofs: shortening proofs (golfing), removing duplication, reorganizing imports, generalizing hypotheses, and optimizing compilation performance; 62\% of these changes span multiple files. At present, this work is mostly performed manually by senior maintainers. In fast-moving blueprint projects, meanwhile, finished proofs may gradually become outdated as mathlib and proof automation evolve, and the room for optimization therefore grows steadily.

\begin{table}[t]
\centering
\footnotesize
\setlength{\tabcolsep}{4pt}
\begin{tabular}{@{}l>{\raggedright\arraybackslash}p{2.5cm}>{\raggedright\arraybackslash}p{2.8cm}@{}}
\toprule
 & \textbf{Instruction-driven} & \textbf{Metric-driven} \\
\midrule
\textbf{Theorem-level} & Automated Theorem Proving, Autoformalization & ImProver, ProofOptimizer, Lean Refactor \\
\textbf{File/repo-level} & APE-Bench, LeanAgent & \textbf{OPAL (this work)} \\
\bottomrule
\end{tabular}
\caption{Positioning of OPAL: repository-level, metric-driven proof optimization is a vacant quadrant.}
\label{tab:positioning}
\end{table}

Automated proof optimization is not a new topic, yet existing work fails to cover this workload along two axes (Table~\ref{tab:positioning}). In terms of scope, proof optimization research has focused on the theorem level: rewriting an individual proof to make it more favorable, typically shorter or faster to compile~\cite{improver,improver2,proofoptimizer,leanrefactor}. In terms of task form, existing proof-engineering benchmarks follow the paradigm of SWE-style coding-agent tasks~\cite{swebench} and are instruction-driven: reproduce one human change given a natural-language instruction, judged by unit or compilation tests together with LLM judges~\cite{apebench,leanagent}. The bulk of human maintenance falls in the vacant intersection of the two axes, repository-level and metric-driven, where the expert or the agent must discover what to change in order to improve the entire library on measurable metrics.

Beyond agent capability, the obstacle to open-ended metric optimization at repository scale is trust: an agent can delete theorems, weaken statements, or silently introduce additional axioms, all of which spuriously improve the metrics. Such reward hacking~\cite{concreteproblems,rewardhacking} has been observed for coding agents on repository-optimization tasks such as GSO~\cite{gso}. Our key observation is that the blueprint structure already declares what must not change: the declarations marked by \texttt{\textbackslash lean\{\}} links are the project's final targets, in practice 5--19\% of all declarations, and their statements must be preserved up to $\alpha$-equivalence; the remaining 81--95\% of auxiliary code (the concrete form of lemmas, the file organization, the proof strategy) is exactly what maintainers routinely refactor and should remain free to change. We formalize this observation as an executable semantic-preservation contract.

Around this contract we build an evaluation environment with a single trust boundary. The agent iterates in a Docker sandbox with advisory self-measurement and profiling tools, while the contract checks and the final metric accounting run outside the sandbox, so the agent cannot forge gains by tampering with its self-measurement tools. OPAL-Bench collects 162 task instances from 18 real-world Lean blueprint projects, covering six optimization objectives mined from mathlib maintenance (proof length, import closure, elaboration cost, duplication of mathlib lemmas, references to deprecated APIs, and blueprint-link deviations), and uses the gains of human historical commits at the corresponding locations as a difficulty reference for each instance.

On OPAL-Bench, a frontier coding agent (Claude Opus 4.8) produces contract-passing improvements on the metric each instance targets for 75.3\% of instances (122/162), at a median of $9.6\times$ the improvement of the human reference commit on the mined subset; the ratio measures the gap between one incremental human commit and an unconstrained sweep of the same region, not per-change capability. The ablation then separates two effects. With the contract absent from the agent's world, 48\% of accepted episodes fail semantic verification, most often because the agent deletes the mathematics being measured; we call these \emph{fabricated gains}. Showing the rules without enforcing them already cuts fabrication to 7\%, so disclosure does most of the work on behavior. Enforcement is still needed, since the remaining fabricated gains look genuine and only external verification can tell them apart.

Our contributions are threefold:
\begin{itemize}
    \item \textbf{Task and benchmark.} A repository-level, metric-driven proof-optimization task whose six objectives are grounded in a large-scale empirical study of maintenance work in mathlib (33{,}222 commits), together with OPAL-Bench, 162 instances from 18 projects with human-headroom normalization.
    \item \textbf{Contract and environment.} An executable semantic-preservation contract that prevents reward hacking by anchoring trust to blueprint structure, together with a taxonomy of cheating behaviors and a sandboxed environment that separates advisory feedback from authoritative verification.
    \item \textbf{Findings.} A frontier coding agent produces contract-passing optimizations beyond the scale of single human commits; ablations show that without the contract nearly half of accepted gains are fabricated, and that disclosing the rules alone prevents most fabrication.
\end{itemize}

\section{Related Work}
\label{sec:related}

\paragraph{Theorem-level proof optimization.}
Automated optimization of existing proofs originates at the theorem level. ImProver first cast proof rewriting as a metric-driven task: a training-free language-model agent, guided by symbolic feedback from Lean, rewrites individual proofs toward user-specified objectives such as length and readability~\cite{improver}. ImProver~2 makes the same task learnable, using expert iteration over a neurosymbolic scaffold to train a small model that matches far larger ones~\cite{improver2}. ProofOptimizer trains a model to compress proofs, measuring proof length by syntax-aware token counting~\cite{proofoptimizer}. Lean Refactor broadens the objective to multiple criteria (length, compilation time, version compatibility), combining a strategy library mined from human history with retrieval and an agent loop, and evaluates on research-level repositories including FLT and PFR~\cite{leanrefactor}. Throughout this line, however, the task formulation remains theorem-level: it cannot express cross-file changes (62\% of human maintenance commits; Section~\ref{sec:study}) or objectives that do not decompose theorem-wise, such as import closure and whole-repository elaboration cost.

\paragraph{Proof engineering in Lean.}
Beyond single-theorem tasks such as automated theorem proving~\cite{gptf,liu2023fimo,liu2026discover} and autoformalization~\cite{autoformalization,xie2025fmc}, proof engineering research has begun to address library maintenance itself. APE-Bench is the first file-level proof engineering benchmark: given a natural-language instruction, the agent must reproduce a human edit, scored by an LLM judge along semantic correctness, requirement alignment, and scope control~\cite{apebench}. LeanAgent brings evaluation to the repository level and submits pull requests to real projects, but its task is completing missing proofs (filling \texttt{sorry}s)~\cite{leanagent}. MathlibPR contributes an instructive negative result: asked to judge whether build-passing mathlib pull requests are merge-ready, state-of-the-art models achieve balanced accuracy no better than chance~\cite{mathlibpr}.

\paragraph{Repository-level optimization in general software.}
In general software engineering, repository-level metric-driven optimization benchmarks have recently proliferated~\cite{gso,sweperf,swefficiency,swerefactor,perfoptbench}, but a shared challenge is measurement reliability: wall-clock-based metrics are machine-dependent, and reference patches satisfy their original validity rules on only a fraction of replayed machines~\cite{perfbench-reliability}. Environment builders such as SWE-smith, SWE-Gym, R2E-Gym, and SWE-rebench automatically turn real repositories into executable agent tasks at scale~\cite{swesmith,swegym,r2egym,swerebench}. They target Python and judge success by unit tests, which check behavior rather than the preservation of what a project has proved. More broadly, in general software both correctness and performance can only be approximated, by generated tests and by wall-clock time, respectively.

OPAL occupies the intersection these lines leave vacant. Its objective functions are defined on the repository as a whole; the agent decides for itself what to change under measurable metrics, rather than reproducing an instructed edit; and trust rests on symbolic verification: semantic preservation is decided by kernel-level machine evidence instead of an LLM judge~\cite{ospanov2026can} or a test distribution, and performance is measured by a deterministic elaboration count instead of wall time.

\section{What Do Mathematicians Actually Optimize?}
\label{sec:study}

Any optimization task embeds a judgment of what counts as \emph{better}. Existing work stipulates it (shorter proofs, faster compilation) without examining what the maintainers of formal libraries actually choose to improve. To ground OPAL empirically, we mine all 33{,}222 commits of mathlib through July 2026, taking as the main analysis window the stable period after the Lean~3-to-4 port~\cite{mathport} (from August 2023, $n=27{,}594$) to exclude porting noise. Since mathlib almost universally follows conventional commits~\cite{conventionalcommits}, we filter commits by their type prefix and label the engineering commits (\texttt{chore}/\texttt{refactor}/\texttt{perf}) with six maintenance tendencies via keyword rules. Blind LLM validation~\cite{llmjudge} places rule precision at 92.5\%; full methodology and the validation protocol appear in Appendix~\ref{app:mining}.

\begin{figure}[t]
\centering
\includegraphics[width=\columnwidth]{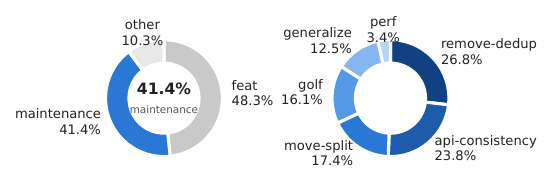}
\caption{Post-port mathlib commits ($n=27{,}594$). Left: overall composition; maintenance commits (\texttt{chore}/\texttt{refactor}/\texttt{perf}) account for 41.4\%, and 62.1\% of them span multiple files. Right: breakdown of maintenance work into the six tendencies, as shares of tendency labels (commits may carry multiple labels).}
\label{fig:composition}
\end{figure}

Maintenance accounts for 41.4\% of commits (11{,}434/27{,}594; Figure~\ref{fig:composition}), and 62.1\% of it spans multiple files, which is invisible from a single-theorem perspective. Move-split commits touch a median of six files.

\begin{table}[t]
\centering
\footnotesize
\setlength{\tabcolsep}{4.5pt}
\begin{tabular}{@{}lrrrrr@{}}
\toprule
\textbf{Tendency} & $n$ & \textbf{Cross-file} & \textbf{Add/Del} & \textbf{Net} & \textbf{Files} \\
\midrule
remove-dedup & 2{,}111 & 56.9\% & $+6/-13$ & $-3$ & 2 \\
api-consistency & 1{,}875 & 72.7\% & $+24/-18$ & $+3$ & 3 \\
move-split & 1{,}373 & 92.5\% & $+89/-66$ & $+17$ & 6 \\
golf & 1{,}271 & 47.0\% & $+8/-18$ & $-7$ & 1 \\
generalize & 984 & 58.8\% & $+34/-18$ & $+7$ & 2 \\
perf & 266 & 44.0\% & $+11/-6$ & $+2$ & 1 \\
\bottomrule
\end{tabular}
\caption{The six maintenance tendencies in post-port mathlib: frequency, share of multi-file commits, and median diff shape (added/deleted lines, net lines, files touched). Commits may carry multiple labels.}
\label{tab:tendencies}
\end{table}

Table~\ref{tab:tendencies} reports the frequency and diff shape of each tendency: \textbf{golf} shortens proofs via stronger automation or newer API (Listing~\ref{lst:golf}); \textbf{move-split} reorganizes files and the import graph; \textbf{remove-dedup} eliminates duplicate lemmas and unused code; \textbf{generalize} weakens hypotheses; \textbf{api-consistency} aligns naming and manages deprecation; and \textbf{perf} optimizes compilation performance, tracked by a continuous benchmarking service~\cite{speedcenter}.

\begin{listing}[t]
{\setlength{\fboxsep}{0pt}\footnotesize\ttfamily\noindent
\newcommand{\diffline}[2]{\colorbox{#1}{\makebox[\columnwidth][l]{\strut #2}}}%
\diffline{diffdel}{- obtain $\langle$p, hp$\rangle$ := p}\\
\diffline{diffdel}{- obtain $\langle$q, hq$\rangle$ := q}\\
\diffline{diffdel}{- rw [Subtype.mk.injEq]}\\
\diffline{diffdel}{- induction p with}\\
\diffline{diffdel}{- ...~~-{}- 21 further lines of induction}\\
\diffline{diffadd}{+ have := p.isPath.exists\_isCycle\_of\_ne}\\
\diffline{diffadd}{+~~~~~~~~~~~q.isPath}\\
\diffline{diffadd}{+ grind [IsAcyclic, Subtype.coe\_inj]}\par}
\caption{A representative golf commit (mathlib, \texttt{IsAcyclic.path\_unique}): a 25-line manual induction proof rewritten in two lines using automation and lemmas added to the library after the original proof was written.}
\label{lst:golf}
\end{listing}

Running the same classification pipeline on three representative blueprint projects (pfr, carleson, sphere-eversion)~\cite{pfr,carleson,sphereeversion} confirms that maintenance activity accounts for a comparable share of commits, with one telling exception: performance-oriented commits, already only 1.0\% of mathlib commits, fall to 0.3\% or less in every blueprint project, leaving elaboration cost an essentially unmaintained dimension. However, blueprint projects differ from mathlib in three structural ways that reshape how each tendency manifests: they typically serve a small set of target theorems rather than a broad API, they rarely have downstream consumers, and they sit on top of a rapidly evolving mathlib whose API changes continuously propagate inward. Three tendencies yield direct metrics: golf maps to proof token count~\cite{proofoptimizer}, move-split to import closure, and perf to elaboration cost (measured by \emph{heartbeats}). Two take blueprint-specific forms, duplication against mathlib and references to deprecated mathlib APIs. A sixth metric captures deviations between the \LaTeX{} plan and the Lean codebase. We collect the six into a metric vector $R$ (defined in Section~\ref{sec:task}), so human maintenance history can be read as incremental improvement on $R$.

\section{Task Definition and the Semantic-Preservation Contract}
\label{sec:task}

Given a Lean blueprint project at a baseline snapshot $P_0$ and the six repository-level metrics $R$ derived in Section~\ref{sec:study}, an agent may apply arbitrary cross-file code changes (moving, rewriting, and reorganizing declarations, and removing mathlib duplicates) to produce a modified project $P'$. The objective is to improve $R(P')$ relative to $R(P_0)$ while satisfying a semantic-preservation contract $C(P_0, P')$. Each task instance specifies a primary objective (one of the six components of $R$) and may optionally provide a scope hint, but no step-by-step instructions; the agent autonomously decides where and how to optimize. Evaluation is determined entirely by the change in $R$ and whether $C$ holds.

All six components of $R$ are measured on the project's own modules, and lower values are better. \texttt{code\_tokens} counts lexical tokens in the project source after comments and docstrings are stripped, so it is insensitive to naming length and layout. \texttt{elab\_heartbeats} sums the elaboration heartbeats of all project modules, a deterministic proxy for compilation cost. \texttt{import\_closure} sums, over local modules, the number of modules in each module's transitive import closure. \texttt{mathlib\_dup} counts local declarations whose elaborated statement is $\alpha$-equivalent to a theorem of an external dependency. \texttt{deprecated\_refs} counts distinct pairs of a local declaration and a deprecated constant it reaches. \texttt{blueprint\_dev} counts blueprint nodes without a \texttt{\textbackslash lean\{\}} link plus links to nonexistent declarations.

Compilation alone is far from sufficient for optimization tasks, and no existing verification method~\cite{lean4checker,proofrefactor} covers a repository-level contract that permits structural changes while preserving semantic content. Such a contract must determine which declarations carry semantic commitments. Verifying only the project's root theorem is insufficient, because blueprint projects routinely contain \texttt{sorry}-truncated proof chains that make the root vacuously satisfied; semantic preservation must be checked at the granularity of individual plan nodes.

Our observation is that blueprint structure itself provides these anchors. The \texttt{\textbackslash lean\{\}} macro in Lean blueprint~\cite{leanblueprint} establishes an explicit mapping between each informal plan node and its corresponding Lean declaration. We parameterize the contract by an anchor set $A$: the type of each declaration in $A$ must be preserved between $P_0$ and $P'$, while proofs of anchored declarations and all non-anchored auxiliary code may be freely modified or extended. In blueprint projects, $A$ consists of the \texttt{\textbackslash lean\{\}} links that resolve to declarations defined in the project itself; links resolving to mathlib are exempt, as they represent content already upstreamed to the library. Anchor sets are small in practice, ranging from 4.7\% to 18.8\% of local declarations across four measured projects; the remaining declarations constitute the refactoring space. The semantic guarantee is therefore scoped to the anchor interface. The contract certifies that anchored statements are preserved, while auxiliary declarations may change their statements but may not disappear. A deletion audit compares the declaration sets of $P_0$ and $P'$ by name and fails any patch that removes a declaration the task does not declare removable. Only \texttt{mathlib\_dup} tasks declare declarations removable, namely the measured duplicates, since deleting them is the task itself. A companion flag fails patches that remove blueprint nodes. Because matching is by name, renaming a declaration counts as a deletion. The anchor-set design is frontend-agnostic: Lean blueprint's \texttt{\textbackslash lean\{\}}, LeanArchitect's \texttt{@[blueprint]} attribute~\cite{leanarchitect}, and Verso's blueprint annotations~\cite{verso} all provide isomorphic node-to-declaration mappings, so the same contract applies across current blueprint frontends.

The contract $C$ enforces semantic preservation of $A$ through the eight verdict rules in Table~\ref{tab:verdict-rules}, all of which a patch must satisfy; statement fingerprints are $\alpha$-equivalence hashes of elaborated expressions. Adversarial testing during development discovered and blocked over ten cheating channels. Design details appear in Appendix~\ref{app:contract}.

\begin{table}[t]
\centering
\footnotesize
\setlength{\tabcolsep}{3pt}
\begin{tabular}{@{}c>{\raggedright}p{2.3cm}>{\raggedright\arraybackslash}p{4.5cm}@{}}
\toprule
\textbf{\#} & \textbf{Check} & \textbf{Prevents} \\
\midrule
1 & Anchor existence & Deletion of anchored declarations \\
2 & Statement fingerprint & Modification of anchored types \\
3 & Definition closure & Semantic drift via redefinition of referenced \texttt{def}s (definition shadowing) \\
4 & Axiom \& \texttt{sorry} count & \texttt{sorry}/axiom injection; per-anchor axiom sets may only shrink \\
5 & Dependency pinning, resource options, docstrings & Toolchain/manifest bumping; \texttt{maxHeartbeats} inflation; anchor docstring deletion \\
6 & Build coverage & Removing modules from build targets \\
7 & Kernel replay (\texttt{leanchecker}) & Environment hacking via metaprogramming \\
8 & Compilation (\texttt{lake build}) & Type-correctness \\
\bottomrule
\end{tabular}
\caption{The eight verdict rules of the semantic-preservation contract.}
\label{tab:verdict-rules}
\end{table}

\begin{figure*}[t]
\centering
\includegraphics[width=\textwidth]{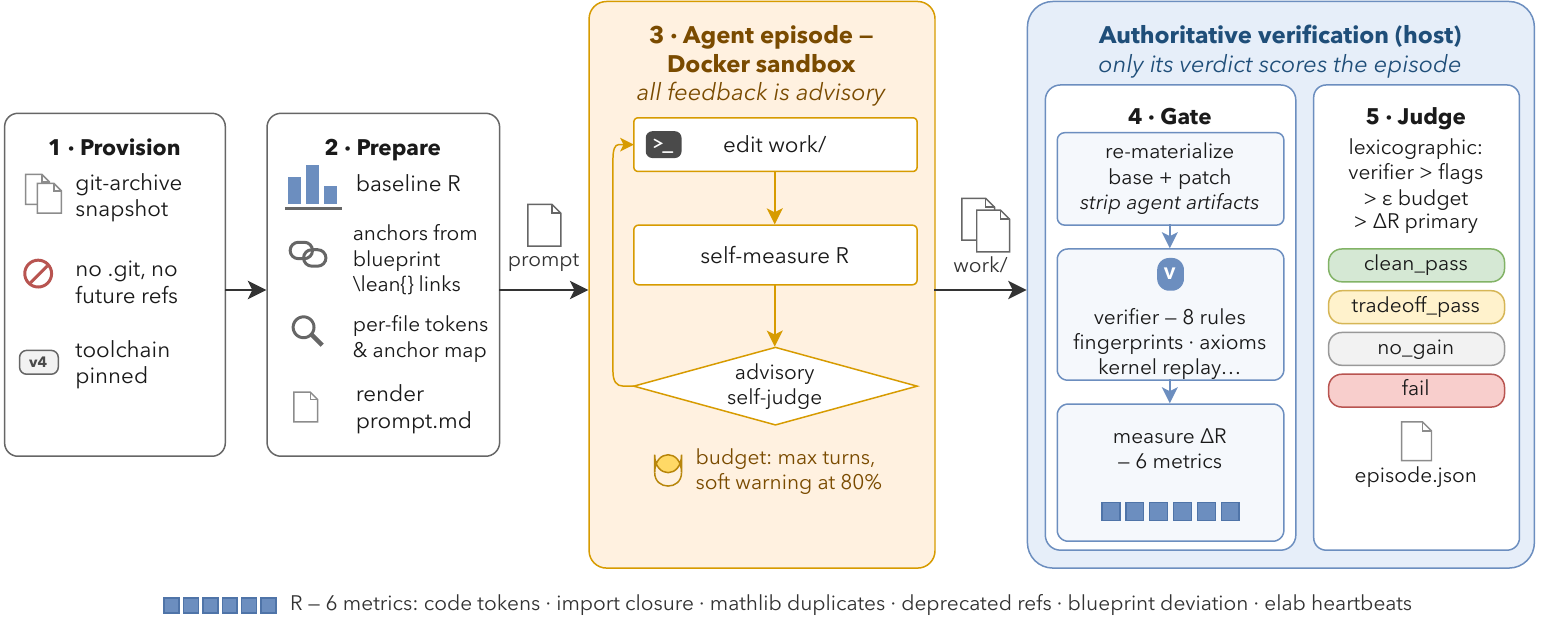}
\caption{The OPAL environment. Host-side setup stages (gray) provision a pristine snapshot, measure the baseline, and extract the anchor set from the blueprint's \texttt{\textbackslash lean\{\}} links. The agent works in a Docker sandbox where all self-measurement is advisory, and its working tree is the only artifact that crosses the trust boundary. The gate then checks the contract and re-measures the six metrics of $R$ (bottom legend) outside the sandbox.}
\label{fig:harness}
\end{figure*}

\paragraph{Evaluation environment and verdicts.}
Each task instance is evaluated in an isolated episode (Figure~\ref{fig:harness}). A coding agent operates inside a Docker sandbox; all self-measurement inside the sandbox is \emph{advisory} and does not enter the score. This separation is what makes the reported numbers trustworthy: the agent gets fast local feedback, while every scored quantity is recomputed by machinery it cannot reach. The episode ends when the agent finishes voluntarily or exhausts its budget. Then authoritative verification runs outside the sandbox: the gate re-materializes both the baseline and the patched project from trusted sources, checks the semantic-preservation contract, and re-measures all six metrics. A lexicographic acceptance policy then issues one of four verdicts: any contract violation yields \verb|fail|; a contract-clean submission with no metric improvement yields \verb|no_gain|; a primary-metric improvement with side regressions within per-component $\varepsilon$ budgets yields \verb|tradeoff_pass|; and an improvement with no side regressions yields \verb|clean_pass|. The policy is a pure function of the gate outputs and a declarative configuration, so strictness variants run as ablation conditions without changing the pipeline. Full environment details appear in Appendix~\ref{app:env}.

\section{OPAL-Bench}
\label{sec:bench}

OPAL-Bench consists of 162 task instances drawn from 18 real-world Lean blueprint projects, selected from public repositories that use the Lean blueprint workflow and build on a recent toolchain (selection criteria in Appendix~\ref{app:bench_projects_selection}), spanning analytic number theory, probability, differential topology, additive combinatorics, harmonic analysis, and other areas of mathematics. The benchmark covers all six optimization objectives and offers three guidance levels. A deterministic 30-instance subset, OPAL-Bench Lite, is provided for rapid ablations and budget profiling (selection procedure in Appendix~\ref{app:bench_lite}).

Instances are constructed from two complementary sources (details in Appendix~\ref{app:bench_sources}). \emph{Historical mining} (56 instances) extracts human optimization commits from project histories; each instance starts from the parent commit, so the agent never sees the human solution, and the human $\Delta R$ serves as a difficulty reference via headroom normalization ($\mathit{ratio} = \Delta R_{\mathit{agent}} / \Delta R_{\mathit{human}}$). \emph{In-the-wild debt} (106 instances) measures each project's current \texttt{HEAD} for discrete debt types (cross-library duplicates, blueprint-link deviations, deprecated API references) and continuous objectives; since no solution exists in any prior corpus, contamination is ruled out by construction. Each discrete debt type yields a pair of instances, one \textbf{exact} (locations listed) and one \textbf{open} (only the debt type stated), isolating discovery ability from elimination ability.

Each instance also carries a guidance level: \textbf{exact} (29 instances) names specific locations, \textbf{module} (56) identifies target files, and \textbf{open} (77) gives only the objective.

\subsection{Scoring}

A contract violation (any of the eight rules in Section~\ref{sec:task}) fails the episode outright with a score of zero; the contract is a hard gate, not a penalty term. For episodes that pass, the reported quantity is the relative change in the primary metric; historically mined instances additionally report the headroom ratio defined above. Results are reported per objective and per source without aggregation into a single scalar.

\subsection{Composition}

Table~\ref{tab:bench_objectives} shows the composition by objective and source; a full per-project breakdown appears in Table~\ref{tab:bench_projects} (Appendix~\ref{app:bench_funnel}).

\begin{table}[t]
\centering
\footnotesize
\setlength{\tabcolsep}{5pt}
\begin{tabular}{@{}lrrr@{}}
\toprule
\textbf{Objective} & \textbf{Hist.} & \textbf{Wild} & \textbf{Total} \\
\midrule
\texttt{code\_tokens} & 50 & 16 & 66 \\
\texttt{mathlib\_dup} & 6 & 30 & 36 \\
\texttt{blueprint\_dev} & 0 & 26 & 26 \\
\texttt{elab\_heartbeats} & 0 & 16 & 16 \\
\texttt{import\_closure} & 0 & 16 & 16 \\
\texttt{deprecated\_refs} & 0 & 2 & 2 \\
\midrule
\textbf{Total} & \textbf{56} & \textbf{106} & \textbf{162} \\
\bottomrule
\end{tabular}
\caption{OPAL-Bench instances by objective and source (Hist.\ = historical mining, Wild = in-the-wild debt). Historical instances concentrate on proof-token golfing, reflecting the distribution of human maintenance activity; in-the-wild instances cover all six objectives.}
\label{tab:bench_objectives}
\end{table}

\section{Experiments}
\label{sec:experiments}

\paragraph{Setup.}
We evaluate all 162 OPAL-Bench instances using Claude Opus~4.8 as the agent backbone, driven by the Claude Agent SDK inside the OPAL environment. Each episode is governed by two hard budget caps: 250 agent interaction turns and a 6-hour wall-clock limit; upon reaching either limit, the current working tree is submitted for evaluation. All verdicts are produced by the contract checker with canonicalized compiler-generated names (Section~\ref{sec:contract_catch}). The median agent cost is \$6.2 per episode (90th percentile \$41.7); the batch totals more than \$2{,}640.

\subsection{Main Results}
\label{sec:main_results}

Table~\ref{tab:main_results} reports outcomes per optimization objective. Across the 162 evaluated instances, the agent produces a contract-passing optimization in 75.3\% of cases (122/162): 100 clean passes (no side regressions) and 22 tradeoff passes (primary metric improved, side regressions within budget). Of the remainder, 34 (21.0\%) fail the contract, an integrity check, or an $\varepsilon$ budget, and 6 (3.7\%) end in no-gain.

\begin{table}[t]
\centering
\footnotesize
\setlength{\tabcolsep}{4pt}
\begin{tabular}{@{}lrrrrr@{}}
\toprule
\textbf{Objective} & $n$ & \textbf{Clean} & \textbf{Trade} & \textbf{Pass\%} & \textbf{Med.~$\Delta R$} \\
\midrule
\texttt{mathlib\_dup}       & 36 & 29 &  3 & 89\% & $-$100\%  \\
\texttt{deprecated\_refs}   &  2 &  2 &  0 & 100\% & $-$100\% \\
\texttt{elab\_heartbeats}   & 16 &  5 &  9 & 88\% & $-$4.1\%  \\
\texttt{blueprint\_dev}     & 26 & 16 &  5 & 81\% & $-$35\%   \\
\texttt{code\_tokens}       & 66 & 42 &  3 & 68\% & $-$1.1\%  \\
\texttt{import\_closure}    & 16 &  6 &  2 & 50\% & $-$3.6\%  \\
\midrule
\textbf{All}                & 162 & 100 & 22 & 75.3\% & ---     \\
\bottomrule
\end{tabular}
\caption{Outcomes by optimization objective. \emph{Clean}: primary metric improved with no side regressions. \emph{Trade}: primary metric improved with side regressions within $\varepsilon$ budgets. \emph{Med.~$\Delta R$}: median relative improvement on the primary metric among passing episodes. For discrete-debt objectives (\texttt{mathlib\_dup}, \texttt{deprecated\_refs}), $-$100\% indicates the agent eliminated the entire identified debt inventory.}
\label{tab:main_results}
\end{table}

\begin{figure}[t]
\centering
\includegraphics[width=\columnwidth]{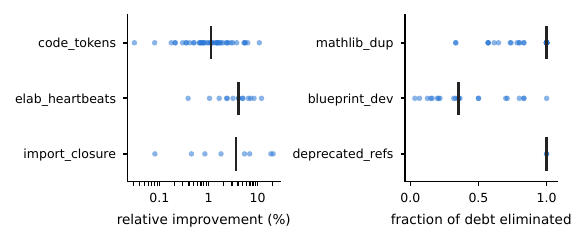}
\caption{Per-episode improvement among passing episodes. Left: relative improvement on continuous objectives (log scale). Right: fraction of the debt inventory eliminated on discrete objectives. Vertical bars mark medians.}
\label{fig:improvement_dist}
\end{figure}

\paragraph{Improvement magnitude.}
A pass requires only strict improvement, so a $-0.03\%$ episode and a $-10.8\%$ episode count equally toward Table~\ref{tab:main_results}; Figure~\ref{fig:improvement_dist} shows what the pass rate conceals. The discrete objectives split into two regimes. On \texttt{mathlib\_dup}, passing episodes eliminate the median inventory completely (18 of 32 clear everything): once a duplicate pair is identified, consolidating it is within the agent's reach. On \texttt{blueprint\_dev} the median passing episode discharges only 35\% of the inventory: blueprint-deviation inventories mix dangling links that maintenance can repair with links whose repair requires new formalization rather than refactoring.

\paragraph{Guidance, sources, and projects.}
Exact instances pass at 86\% versus 74\% for open; historical and in-the-wild sources agree (71\% vs.\ 77\%), so the conclusions do not rest on the contamination-exposed source. Project-level variance is large: PrimeNumberTheoremAnd, the largest project, passes only 3 of 11 instances. Voluntary completions (77\% of episodes) pass at 86\%, while budget-capped episodes pass at only 39\% (turn cap) and 44\% (wall-clock cap). The headline numbers are a property of the frontier agent: two open-weight models on the Lite subset convert far fewer instances (Appendix~\ref{app:crossmodel}), but the verification layer transfers unchanged across models.

\subsection{What Does the Contract Catch?}
\label{sec:contract_catch}

Of the 34 failed episodes, Table~\ref{tab:contract_catch} breaks down which rules fired.

\begin{table}[t]
\centering
\footnotesize
\setlength{\tabcolsep}{4pt}
\begin{tabular}{@{}llr@{}}
\toprule
\textbf{Rule / Flag} & \textbf{Violation type} & $n$ \\
\midrule
Rule 8  & Build failure               & 12 \\
Rule 3  & Definition-closure change   & 9  \\
Rule 2  & Anchor statement change     & 4  \\
Rule 7  & Kernel replay failure       & 3  \\
Rule 4  & Axiom-set regression        & 3  \\
Rule 6  & Module dropped from build   & 1  \\
Rule 5  & Resource-option guard       & 1  \\
\midrule
Deletion audit & Undeclared deletions & 7  \\
$\varepsilon$ & Heartbeat regression  & 1  \\
\bottomrule
\end{tabular}
\caption{Rules fired across the 34 failed episodes under the canonicalized checker. An episode may trigger several rules, so counts sum to more than 34.}
\label{tab:contract_catch}
\end{table}

\paragraph{Failures of self-assessment, not of intent.}
The dominant pattern is that the agent does not know it has failed: most rule-2/3 violations and several build failures end with the agent finishing voluntarily, convinced it produced a valid optimization. Drift arises from ordinary agent overconfidence rather than adversarial pressure.

\paragraph{Metric gaming concentrates where legitimate optimization is hardest.}
Rule~4, the per-anchor axiom audit, fires three times, all on PrimeNumberTheoremAnd, the hardest project. In one episode the agent replaced a proof body with \texttt{sorry}, making elaboration nearly free while the theorem ceases to be proved; the other two regressed anchored declarations' axiom sets. That gaming concentrates where honest gains are hardest suggests a pressure-dependent failure mode; the ablation in Section~\ref{sec:ablation_contract} tests whether removing the contract makes it the default. A detailed audit appears in Appendix~\ref{app:gaming_audit}.

\subsection{Ablation: What Does the Contract Buy?}
\label{sec:ablation_contract}

We run three conditions on the 30-instance Lite subset (Section~\ref{sec:bench}), holding the model, budget, and task pool fixed and varying only what the agent is told and what the scorer enforces: \textbf{Full} (the main-results configuration, contract enforced), \textbf{Advisory} (rules shown but not enforced; acceptance by primary metric only), and \textbf{None} (contract absent from prompt and sandbox; acceptance by primary metric only). Every episode from all three conditions is re-scored offline under the full contract. An episode that its own condition accepted but the full contract rejects is a \emph{fabricated gain} (Table~\ref{tab:ablation_contract}).

\begin{table}[t]
\centering
\footnotesize
\setlength{\tabcolsep}{4pt}
\begin{tabular}{@{}lrrrr@{}}
\toprule
\textbf{Condition} & \textbf{Accepted} & \textbf{Fabricated} & \textbf{Genuine} & \textbf{Med.~$\Delta R$} \\
\midrule
Full      & 26/30 & 0 (0\%)            & 26 & $-$20.3\% \\
Advisory  & 27/30 & 2 (7\%)            & 25 & $-$12.5\% \\
None      & 27/30 & 13 (48\%)          & 14 & $-$32.6\% \\
\bottomrule
\end{tabular}
\caption{Contract ablation on OPAL-Bench Lite. \emph{Accepted}: episodes the condition's own policy accepts (for Full, the Lite subset of the main batch, this means passing the contract, so fabrication is zero by construction). \emph{Fabricated}: of those, the count the full contract rejects under offline re-scoring. \emph{Genuine}: episodes accepted by both. \emph{Med.~$\Delta R$}: median relative improvement among genuine passes.}
\label{tab:ablation_contract}
\end{table}

\paragraph{Fabricated-gain rates.}
Offline re-scoring rejects 13 of the None condition's 27 accepted episodes (48\%), leaving a genuine yield of 14/30 against 26/30 for Full, so the unverified success rate overstates real capability almost twofold. Merely \emph{showing} the rules recovers most of the gap (Advisory fabricates 2 of 27), so an agent told what must be preserved largely preserves it.
\paragraph{How fabrication manifests.}
The 13 fabricated episodes are dominated by deletion of content (declarations, modules, or proof bodies replaced by \texttt{sorry}), the shortest path to a size metric. Their claimed gains look ordinary, so only verification separates them from genuine ones. Detailed violation profiles appear in Appendix~\ref{app:fabricated_profiles}.

\subsection{Ablation: Which Environment Components Matter?}
\label{sec:ablation_env}

We decompose the environment into four cumulative levels on the Lite subset, from bare (no metric tooling) through profiling, feedback, to the full contract, and re-judge all conditions under the full contract for comparability (Table~\ref{tab:ablation_env}). Pass rates barely respond to instrumentation (40\% bare, 33\% profiling, 53\% feedback) but jump to 87\% when the contract is added; the rejection rate drops from roughly half to 13\%. The environment's contribution to \emph{validity} is concentrated in governance, not instrumentation. Level definitions and detailed analysis appear in Appendix~\ref{app:env_ablation}.

\begin{table}[t]
\centering
\footnotesize
\setlength{\tabcolsep}{4pt}
\begin{tabular}{@{}lrrrr@{}}
\toprule
\textbf{Level} & \textbf{Pass\%} & \textbf{Med.~$\Delta R$} & \textbf{Rejected} & \textbf{Turns} \\
\midrule
L0 bare        & 40\% & $-$20.3\% & 57\% & 116 \\
L1 +~profiling & 33\% & $-$14.8\% & 63\% & 122 \\
L2 +~feedback  & 53\% & $-$20.8\% & 40\% & 118 \\
L3 full        & 87\% & $-$20.3\% & 13\% & 142 \\
\bottomrule
\end{tabular}
\caption{Environment-component ablation on OPAL-Bench Lite, all conditions judged under the full contract. \emph{Rejected}: share of episodes the full contract rejects. \emph{Turns}: median agent turns consumed.}
\label{tab:ablation_env}
\end{table}

\subsection{Baseline: Is Repository Scope Necessary?}
\label{sec:baseline_theorem}

We compare the repository-level agent against a theorem-level baseline that may only rewrite proof bodies, with all other variables held fixed on the Lite subset. Of the 30 instances, 17 target cross-file debt that proof-body edits cannot express by construction. On the remaining 13, the theorem-level baseline passes 7 against the repository-level configuration's 10; where both pass, the constrained agent captures a smaller fraction of the available gain. Over the full Lite inventory, the theorem-level ceiling is 7/30 against 26/30: most of what a repository-level optimizer is worth lies in debt that a theorem-level system cannot express. Per-instance details appear in Appendix~\ref{app:baseline_theorem}. Appendix~\ref{app:experiments} further reports discovery versus execution on exact/open pairs, headroom over human commits, cross-objective interactions, and case studies.

\section{Conclusion}
\label{sec:conclusion}

We introduced OPAL, a repository-level, metric-driven proof-optimization task grounded in what maintainers of formal libraries actually optimize. Its central observation is that blueprint structure already supplies the trust boundary such a task needs, since a contract over the 5--19\% of declarations pinned by \texttt{\textbackslash lean\{\}} links frees the remaining 81--95\% for open-ended refactoring. Around the contract we built an evaluation environment that separates advisory in-sandbox feedback from authoritative external verification, and OPAL-Bench, 162 instances from 18 real-world Lean blueprint projects with human commits as difficulty references.

On OPAL-Bench, a coding agent produces contract-passing optimizations on 75.3\% of instances, while removing the contract from the agent's world makes fabricated gains systematic. The task is far from solved. Agents struggle on the largest projects and long horizons, and often cannot tell when their own patch breaks the contract. OPAL-Bench also separates models clearly, with Claude Opus~4.8 passing 26/30 Lite instances against 14/30 for DeepSeek v4-pro and 5/30 for Kimi k2.7-code (Appendix~\ref{app:crossmodel}), which makes it a useful benchmark for evaluating LLM agents. Formal libraries are a setting where both the objective and its safeguards can rest on machine evidence, with deterministic metrics in place of wall-clock time and kernel-checked semantic preservation in place of generated tests or LLM judges. OPAL shows that under such verifiable conditions, open-ended optimization can be delegated to agents while formally guaranteeing that the anchored statements are preserved.

\section*{Acknowledgements}
This paper is partially supported by the National Key Research and Development Program of China with Grant No. 2023YFC3341203 as well as the National Natural Science Foundation of China with Grant Number 62276002.

\bibliography{aaai2027}

\appendix
\section{Limitations}
\label{sec:limitations}

First, the scale of our experiments is bounded by cost: repository-level optimization is demanding enough that each episode grants the agent 250 turns and up to six hours of wall-clock time, at a median cost of \$6.2, so a single pass over all 162 instances already totals more than \$2{,}640. The full benchmark is therefore evaluated with one rollout per instance, and all ablations run on the 30-instance Lite subset; per-instance variance under repeated rollouts remains uncharacterized.

Second, the contract is a conservative approximation of semantic preservation rather than an exact characterization: it certifies that anchored statements, their definition closures, per-anchor axiom sets, and downstream integrity are preserved, though it rejects some legitimate edits, such as replacing an anchor statement with a logically equivalent or more general formulation. The anchor set itself is inherited from the project's \texttt{\textbackslash lean\{\}} links, whose coverage and granularity reflect the habits of individual repository authors; curating or revising these links, beyond the repair of dangling links measured by \texttt{blueprint\_dev}, lies outside the agent's action space in this work.

\section{Commit Mining Methodology}
\label{app:mining}

This section details the data collection, classification rules, and validation protocol for the empirical study in Section~\ref{sec:study}.

\subsection{Data Collection}

We extract the full commit history of mathlib4 via \texttt{git log --numstat}, recording for each commit the SHA, timestamp, author, subject line, and per-file addition/deletion counts. The resulting dataset contains 33{,}222 commits. Each commit subject is parsed for a conventional-commit prefix (\texttt{feat}, \texttt{chore}, \texttt{refactor}, \texttt{perf}, \texttt{fix}, \texttt{doc}, \texttt{ci}, \texttt{style}, \texttt{test}) and a trailing PR number. The main analysis window covers the post-port period from August 2023 onward ($n=27{,}594$), excluding the Lean~3-to-4 porting era during which commit patterns are dominated by mechanical translation rather than maintenance decisions.

\subsection{Classification Rules}

Engineering commits (those with prefix \texttt{chore}, \texttt{refactor}, or \texttt{perf}) are further labeled with nine subcategory tags via keyword rules applied to the commit subject. The six maintenance tendencies and three noise categories, together with representative patterns, are listed in Table~\ref{tab:rules}. Rules are applied as case-insensitive regular expressions; a commit may receive multiple labels when its subject matches patterns from more than one category.

\begin{table}[h]
\centering
\footnotesize
\setlength{\tabcolsep}{3.5pt}
\begin{tabular}{@{}ll>{\raggedright\arraybackslash}p{4.2cm}@{}}
\toprule
\textbf{Category} & \textbf{Type} & \textbf{Representative patterns} \\
\midrule
golf & tendency & \texttt{golf}, \texttt{simplif*}, \texttt{tidy}, \texttt{shorten}, \texttt{use* \{tactic\}} \\
move-split & tendency & \texttt{move}, \texttt{split}, \texttt{upstream}, \texttt{reorganize}, \texttt{reduce imports} \\
remove-dedup & tendency & \texttt{remove}, \texttt{delete}, \texttt{dedup*}, \texttt{duplicat*}, \texttt{unused}, \texttt{redundant} \\
generalize & tendency & \texttt{generalize}, \texttt{weaken}, \texttt{relax}, \texttt{drop *hypothesis} \\
api-consistency & tendency & \texttt{rename}, \texttt{deprecate}, \texttt{deprecating}, \texttt{align}, \texttt{redefin*}, \texttt{notation} \\
perf & tendency & \texttt{speed up}, \texttt{faster}, \texttt{performance}, \texttt{instance priorit*} \\
\midrule
infra & noise & \texttt{bump}, \texttt{nolint*}, \texttt{toolchain}, \texttt{nightly}, \texttt{backport}, \texttt{ci} \\
port-cleanup & noise & \texttt{porting note}, \texttt{mathport} \\
docs-style & noise & \texttt{typo}, \texttt{docstring}, \texttt{whitespace}, \texttt{spelling} \\
\bottomrule
\end{tabular}
\caption{Keyword classification rules for engineering commits. Patterns are abbreviated; the full regular expressions are available in the released code.}
\label{tab:rules}
\end{table}

Commits with the \texttt{perf} prefix automatically receive the \texttt{perf} label even if no keyword matches, since the prefix itself signals a performance-oriented change.

\subsection{Coverage}

The keyword rules assign at least one label to 67.6\% of engineering commits (7{,}732/11{,}434). The remaining 32.4\% consists of semantically implicit subjects (e.g., ``improve notation,'' ``curry \texttt{mk}'') that resist keyword matching; we handle this via coverage disclosure rather than LLM-based bulk labeling, to keep the classification fully reproducible.

\subsection{Precision Validation}

We validate rule precision via blind LLM annotation. From the post-port engineering subset, we draw a stratified sample of 80 commits (10 per label across 8 categories, seed = 42). Eight independent Claude Haiku agents, each seeing only the commit subject and a label definition (not the rule-assigned label), produce multi-label annotations. A rule label is counted as correct if it appears in the annotator's label set.

Overall precision is 92.5\% (74/80). Per-label results are shown in Table~\ref{tab:precision}. Four labels achieve perfect precision; \texttt{remove-dedup} and \texttt{api-consistency} each have three mismatches, primarily on boundary cases involving deprecated declarations (which can be classified as either removal or API work depending on perspective).

\begin{table}[h]
\centering
\footnotesize
\begin{tabular}{@{}lrrr@{}}
\toprule
\textbf{Label} & $n$ & \textbf{Agree} & \textbf{Precision} \\
\midrule
golf & 10 & 10 & 1.00 \\
move-split & 10 & 10 & 1.00 \\
remove-dedup & 10 & 7 & 0.70 \\
generalize & 10 & 10 & 1.00 \\
api-consistency & 10 & 7 & 0.70 \\
perf & 10 & 10 & 1.00 \\
infra & 10 & 10 & 1.00 \\
port-cleanup & 10 & 10 & 1.00 \\
\midrule
\textbf{Overall} & \textbf{80} & \textbf{74} & \textbf{0.925} \\
\bottomrule
\end{tabular}
\caption{Per-label precision of keyword rules under blind LLM validation.}
\label{tab:precision}
\end{table}

\subsection{Unlabeled Tail Composition}

To check whether the unlabeled 32.4\% harbors a systematically missing category, we draw a random sample of 200 unlabeled engineering commits (seed = 42) and have 10 independent Claude Haiku agents classify each by semantic content, with an explicit \texttt{none} option requiring a short topic description.

Of these 200 commits, 171 (85.5\%) are assignable to the existing nine categories, with structural tendencies (api-consistency, golf, move-split, remove-dedup, generalize) again dominating. The remaining 29 (14.5\%) are genuine outliers: typo and docstring fixes, new lemma additions, and test additions that were miscategorized under engineering prefixes. No systematically missing tendency emerges from this analysis.

\section{Semantic-Preservation Contract Details}
\label{app:contract}

This section details the eight verdict rules of the semantic-preservation contract introduced in Section~\ref{sec:task}, along with key design decisions and adversarial validation.

\subsection{Verdict Rules}

Table~\ref{tab:verdict-rules} (Section~\ref{sec:task}) lists the eight rules. A patch from $P_0$ to $P'$ must satisfy all rules to be accepted; only Rule~8 (compilation) triggers early termination.

\subsection{Statement Fingerprinting}

Statement fingerprinting (Rule~2) operates on Lean's \emph{elaborated expressions} (the fully resolved internal representation produced after desugaring, implicit-argument insertion, and type inference) rather than on pretty-printed strings. The fingerprint is an $\alpha$-equivalence hash that normalizes binder names, erases metadata nodes, and positionalizes universe parameters. Pretty-printed strings are unsuitable for this purpose: custom notations can cause semantically distinct statements to print identically, while version-dependent formatting changes can cause semantically equivalent statements to print differently.

\subsection{Definition-Closure Integrity}

The definition-closure check (Rule~3) guards against \emph{definition shadowing}. If an anchored theorem's type references a project-local \texttt{def}, modifying that definition's body leaves the type textually unchanged while altering its meaning through definitional unfolding. Rule~3 prevents this by requiring that the definition bodies of all project-local \texttt{def}s transitively reachable from each anchor's type remain hash-identical between $P_0$ and $P'$. The closure traversal follows \texttt{def}/\texttt{abbrev}/\texttt{instance} value bodies and inductive constructor names, but does not enter theorem proof terms (by proof irrelevance).

\subsection{Resource-Option Guards}

Rule~5 includes guards against resource-limit inflation (\texttt{maxHeartbeats}, \texttt{maxRecDepth}). A violation is raised if either the occurrence count or the maximum value of any resource option increases from $P_0$ to $P'$.

\section{The OPAL Environment}
\label{app:env}

Figure~\ref{fig:harness} shows the evaluation pipeline. Each task instance is processed as an isolated \emph{episode} through the five stages described below, numbered left to right in the figure; the trust boundary separates the advisory sandbox (stage~3) from authoritative verification (stages~4--5).

\subsection{Stage 1: Provision}

A pristine source snapshot of the project at its pinned commit and toolchain is exported via \texttt{git archive}. The resulting checkout contains no \texttt{.git} directory and no references to commits after the base commit, closing the future-leakage channel by construction.

\subsection{Stage 2: Prepare}

The environment measures the baseline $R(P_0)$, extracts the anchor set from the blueprint's \texttt{\textbackslash lean\{\}} links, and builds a repository profile (per-file token mass and anchor locations). These are rendered, together with the metric definitions, the acceptance rules, and the budget, into the agent's prompt. Listing~\ref{lst:prompt} shows a complete rendered prompt for reference.

\subsection{Stage 3: Agent Episode}
\label{sec:env}

The environment is agent-agnostic: any coding agent that can operate on a directory plugs in through a thin adapter (our experiments use the Claude Agent SDK). Inside a Docker sandbox, the agent edits the working tree, builds the project, and iterates without step-level supervision. The budget is enforced as a hard cap on agent turns, with a single soft warning at 80\% asking the agent to converge to a compiling state. Episodes that end by budget exhaustion are scored on whatever state the working tree is in, so unfinished exploration is not rewarded.

The central design principle is that \emph{everything the agent observes inside the sandbox is advisory}: the agent is given the same measurement and judging tools that the evaluation uses, but none of its self-reported numbers enter the score. The ablations confirm this: agents whose work does not survive verification often finish voluntarily, convinced of success (Sections~\ref{sec:contract_catch} and~\ref{sec:ablation_contract}).

Full measurement of $R$ requires a whole-repository build and per-module re-elaboration (10--20 minutes on benchmark projects), which is too slow for an edit-measure loop. The environment therefore exposes two tiers: a \emph{fast} tier that measures only proof-token count in seconds, and an \emph{env} tier that measures all six metrics.

The agent container runs as the invoking user (not root), with the episode directory as its only writable mount. \texttt{HOME} points into the episode directory, so all agent-side state is sealed per episode. Memory is capped by cgroup with swap disabled, so a runaway build is killed inside the container instead of degrading the host.

\subsection{Stage 4: Gate}

The gate's threat model is that the entire episode directory is agent-controlled. The baseline is re-exported from the trusted snapshot; the patch side is reconstructed from the agent's working tree with all build artifacts removed and rebuilt from source. Contract checking and metric measurement run in single-use containers with no credentials and a fresh \texttt{HOME}; the shared baseline cache is mounted only into baseline-side containers, while the patch side uses a private, per-episode cache. The gate then runs the semantic-preservation contract (Section~\ref{sec:task}) and, if it passes, measures $\Delta R$ across all six metrics.

\subsection{Stage 5: Judge}

A lexicographic acceptance policy converts the gate's outputs into one of four verdicts: any contract violation or integrity flag (the deletion audit and the blueprint-node flag of Section~\ref{sec:task}) yields \verb|fail|; a contract-clean submission with no metric improvement yields \verb|no_gain|; a primary-metric improvement with side regressions within per-component $\varepsilon$ budgets yields \verb|tradeoff_pass|; and an improvement with no side regressions yields \verb|clean_pass|.

\subsection{Agent Prompt}

Listing~\ref{lst:prompt} shows the complete prompt rendered for a \texttt{code\_tokens} (proof-golfing) instance under the main-results configuration (L3). Prompts for other objectives follow the same structure; the goal, measurement commands, intel table, and hard constraints are adapted to the target metric. Template variables (\texttt{\$baseline\_tokens}, \texttt{\$measure\_cmd}, etc.) are filled at episode creation time from the baseline measurement and instance configuration.

\begin{lstlisting}[language={},basicstyle={\scriptsize\ttfamily},numbers=none,xleftmargin=0em,breaklines=true,float=false]
# Task: reduce this repository's code_tokens metric

You are in a sandbox copy of a Lean 4 project (the current
directory), with full read/write and command-execution
permissions.

## Goal

Minimize the `code_tokens` component of the R vector (the
total number of lexical tokens in the source inside the
root_modules import closure, after stripping comments/
docstrings; dead files not imported by any root module do
not count, so stubbing them out or deleting them yields
nothing).
Current baseline: **$baseline_tokens tokens** (baseline
measurement file: `$baseline_json`).

## Scope hint (advisory)

$scope_hint
This hint points at a region known to be optimizable; it
is a suggested entry point, not a hard boundary --
legitimate optimizations outside it count too.

## Measurement and self-assessment tools

- Measure the current state (seconds): `$measure_cmd`
- Compare against the baseline: `$diff_cmd`
- Self-check using the same rules as the external
  authority (advisory): `$policy_cmd`

## Intel (precomputed by the environment, advisory)

Top-8 token distribution inside the root_modules import
closure. The complete per-file list -- including, for each
file, the declaration names anchored to it
(`anchored_decls`) and unlocated anchors
(`unmapped_anchors`) -- is in `$token_report_json`:

$token_report_table

The anchor counts are a textual-heuristic approximation:
before touching a file, check its `anchored_decls` list
name by name. Files with few anchors are less constrained;
consider them first.

## Grading rules

$advisory_notice$policy_summary

External verification runs on a fresh copy outside the
sandbox; none of your in-sandbox self-test results are
taken at face value.

## Hard constraints

1. The project must pass `lake build`;
2. For anchor declarations: name and statement must not
   change, docstrings must not be deleted; proof bodies may
   be freely rewritten; the definition closure reachable
   from an anchor statement is fingerprint-protected;
3. Do not modify lean-toolchain, lake-manifest.json,
   lakefile.toml/lakefile.lean;
4. Do not add or raise maxHeartbeats / maxRecDepth;
5. Do not introduce sorry or new axioms;
6. Do not delete declarations: any disappearing declaration
   trips the undeclared_deletions integrity flag.

## How to work: multi-round expansion + git checkpoints

The workspace is not a git repository. Before starting, run
`git init && git add -A && git commit` to establish a
baseline; after each successful edit-build-measure loop,
`git commit` to bank progress. Committed work cannot be
lost by a later failed attempt.

## Budget

Turn cap: $max_turns. Wall-clock cap: $wall_clock_min min.
At 80% of the turn budget a wrap-up reminder arrives; stop
new exploration, roll back unfinished fronts, and enter the
wrap-up flow.
\end{lstlisting}
\captionof{listing}{Complete rendered prompt for a \texttt{code\_tokens} instance (L3 configuration). Template variables (prefixed \texttt{\$}) are filled from the baseline measurement and instance configuration at episode creation time.}
\label{lst:prompt}

\section{Benchmark Construction Details}
\label{app:bench_funnel}

\subsection{Project Selection}
\label{app:bench_projects_selection}

The 18 benchmark projects were collected by searching GitHub and the Lean community Zulip for public Lean~4 repositories that use the Lean blueprint workflow (i.e., contain a \texttt{blueprint/} directory with \texttt{\textbackslash lean\{\}} links). We required each project to (i)~build successfully on a recent Lean~4 toolchain ($\geq$~v4.28), (ii)~contain at least 500 local declarations (excluding mathlib dependencies) to ensure a non-trivial optimization surface, and (iii)~have an active commit history rather than being a toy or abandoned repository. The resulting pool spans a broad range of mathematical areas and project sizes, from pfr (1{,}303 local declarations) to PrimeNumberTheoremAnd (over 6{,}000).

\subsection{Two Complementary Sources}
\label{app:bench_sources}

\paragraph{Historical mining.}
From over 25{,}000 commits across the project pool, a five-stage funnel (tendency keyword matching, granularity window, toolchain compatibility, measured $\Delta R$ significance, and baseline validity) extracts 56 instances in which a human maintainer demonstrably improved a metric. Each instance takes the \emph{parent} commit of the human optimization as its starting snapshot: the agent is told which metric to target and given a scope hint, but never sees the human solution. The human $\Delta R$ serves as a difficulty reference, enabling \emph{headroom normalization}: the score is $\mathit{ratio} = \Delta R_{\mathit{agent}} / \Delta R_{\mathit{human}}$, so a ratio of 1.0 means the agent matched the human, and raw $\Delta R$ differences across instances of varying scale are placed on a common footing. Leakage is prevented by three measures: \texttt{git archive} expansion (no \texttt{.git}, no refs), a prompt whitelist that structurally excludes the answer commit, and directory-name demasking that replaces the answer SHA with a one-way hash.

\paragraph{In-the-wild debt.}
For each project's current \texttt{HEAD}, a full metric measurement produces a debt inventory: $\alpha$-equivalent duplicate declaration pairs against mathlib, blueprint-link deviations (missing or dangling \texttt{\textbackslash lean\{\}} links), and references to deprecated mathlib APIs. Each discrete debt type yields a \emph{pair} of instances: an \textbf{exact} instance that lists the specific locations (e.g., ``eliminate these five duplicate pairs''), and an \textbf{open} instance that states only the debt type (e.g., ``this project contains duplicates of mathlib declarations''). The paired difference isolates \emph{discovery} ability from \emph{elimination} ability. Continuous objectives (proof-token golfing, import-closure reduction, elaboration-cost reduction) produce only open instances, since no discrete location list exists. This source contributes 106 instances.

\paragraph{Why two sources?}
Historical mining provides human-calibrated difficulty references but carries a training-data contamination risk: the human solutions are public GitHub commits that language models may have seen. In-the-wild debt has no human reference, but the solutions \emph{do not exist} in any prior corpus, so contamination is ruled out by construction.

\subsection{Historical Mining Funnel}

\begin{figure}[t]
\centering
\includegraphics[width=\columnwidth]{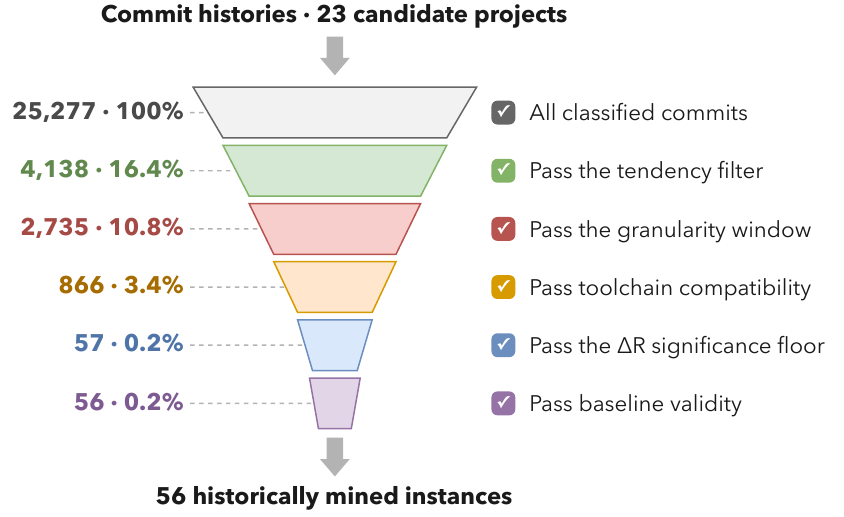}
\caption{The historical mining funnel. Five successive filters reduce 25{,}277 classified commits to 56 benchmark instances; percentages are relative to the starting pool.}
\label{fig:mining_funnel}
\end{figure}

The five-stage funnel that produces historically mined instances (Figure~\ref{fig:mining_funnel}) operates as follows. Starting from 25{,}277 classified commits across the 23-project candidate pool (the 18 benchmark projects are the members of this pool that survive the exclusions below and contribute at least one instance):

\begin{enumerate}
    \item \textbf{Tendency filter.} Retain commits whose subcategory labels match one of the six maintenance tendencies (Section~\ref{sec:study}). Yield: 4{,}138.
    \item \textbf{Granularity window.} Retain commits with churn between 2 and 120 lines and touching at most 4 files, ensuring the change is large enough to produce a measurable signal but small enough to attribute to a single optimization intent. Yield: 2{,}735.
    \item \textbf{Toolchain compatibility.} Retain commits whose parent uses Lean~4 with toolchain $\geq$ v4.28, the minimum version supported by the measurement and contract infrastructure. Yield: 866.
    \item \textbf{$\Delta R$ significance.} Check out parent and child in a git worktree, measure $R$ on both sides, and retain commits where the target metric component improves by at least a pre-set floor. For elaboration heartbeats, the floor is calibrated from repeated measurements (3 projects $\times$ 5 repetitions; floor $= \max(10{,}000,\; 0.001 \times R_0)$, set at $3\times$ the observed maximum run-to-run range). Yield: 57 candidate instances (each significantly improved objective component of a surviving commit produces one candidate).
    \item \textbf{Baseline validity.} Retain candidates whose starting snapshot elaborates without errors, so that baseline metrics are well defined. One candidate is removed: its parent snapshot fails to elaborate (an unknown-identifier error), reproducibly and independently of any agent. Yield: 56 instances.
\end{enumerate}

Two project-level exclusions apply by default: projects flagged as AI-driven (commit language inconsistent with human maintenance) and projects with generated content (distorting $\Delta R$ baselines) are excluded from historical mining.

\begin{table}[h]
\centering
\footnotesize
\setlength{\tabcolsep}{3pt}
\begin{tabular}{@{}lrrrrrr|r@{}}
\toprule
\textbf{Project} & \textbf{G} & \textbf{D} & \textbf{B} & \textbf{P} & \textbf{I} & \textbf{M} & $\Sigma$ \\
\midrule
Noperthedron    & 32 &  2 &  2 & 1 & 1 & 0 & 38 \\
brownian-motion &  8 &  3 &  2 & 1 & 1 & 0 & 15 \\
FLT             &  4 &  6 &  0 & 1 & 1 & 0 & 12 \\
PNT             &  5 &  2 &  2 & 1 & 1 & 0 & 11 \\
carleson        &  3 &  2 &  0 & 1 & 1 & 2 &  9 \\
apap            &  2 &  2 &  2 & 1 & 1 & 0 &  8 \\
BrauerGroup     &  1 &  2 &  2 & 1 & 1 & 0 &  7 \\
CLT             &  1 &  2 &  2 & 1 & 1 & 0 &  7 \\
ClassFieldTheory&  1 &  2 &  2 & 1 & 1 & 0 &  7 \\
DividedPowers4  &  1 &  2 &  2 & 1 & 1 & 0 &  7 \\
LeanModularForms&  1 &  2 &  2 & 1 & 1 & 0 &  7 \\
infinity-cosmos &  1 &  2 &  2 & 1 & 1 & 0 &  7 \\
pfr             &  1 &  2 &  2 & 1 & 1 & 0 &  7 \\
proetale        &  1 &  2 &  2 & 1 & 1 & 0 &  7 \\
sphere-eversion &  1 &  2 &  0 & 1 & 1 & 0 &  5 \\
toric           &  1 &  0 &  2 & 1 & 1 & 0 &  5 \\
flt-regular     &  2 &  0 &  0 & 0 & 0 & 0 &  2 \\
Sphere-Packing  &  0 &  1 &  0 & 0 & 0 & 0 &  1 \\
\midrule
\textbf{Total}  & \textbf{66} & \textbf{36} & \textbf{26} & \textbf{16} & \textbf{16} & \textbf{2} & \textbf{162} \\
\bottomrule
\end{tabular}
\caption{Per-project instance counts.\protect\footnotemark{} G = \texttt{code\_tokens} (golf), D = \texttt{mathlib\_dup} (dedup), B = \texttt{blueprint\_dev}, P = \texttt{elab\_heartbeats} (perf), I = \texttt{import\_closure}, M = \texttt{deprecated\_refs} (migrate).}
\label{tab:bench_projects}
\end{table}
\subsection{OPAL-Bench Lite}
\label{app:bench_lite}

OPAL-Bench Lite is a deterministic 30-instance subset of the full benchmark, designed for rapid ablation experiments and budget profiling. The selection algorithm balances six per-objective quotas to flatten the full benchmark's skew toward proof-token golfing, preserves exact/open pairs (so paired analysis remains possible on the subset), stratifies historically mined instances by headroom magnitude (small/medium/large human $\Delta R$), and caps any single project at three instances (Table~\ref{tab:bench_lite_projects}).

\footnotetext{GitHub repositories, in table row order:\\
\scriptsize
\texttt{github.com/jcreedcmu/Noperthedron}\\
\texttt{github.com/RemyDegenne/brownian-motion}\\
\texttt{github.com/ImperialCollegeLondon/FLT}\\
\texttt{github.com/AlexKontorovich/PrimeNumberTheoremAnd}\\
\texttt{github.com/fpvandoorn/carleson}\\
\texttt{github.com/YaelDillies/apap}\\
\texttt{github.com/Whysoserioushah/BrauerGroup}\\
\texttt{github.com/RemyDegenne/CLT}\\
\texttt{github.com/kbuzzard/ClassFieldTheory}\\
\texttt{github.com/AntoineChambert-Loir/DividedPowers4}\\
\texttt{github.com/CBirkbeck/LeanModularForms}\\
\texttt{github.com/emilyriehl/infinity-cosmos}\\
\texttt{github.com/teorth/pfr}\\
\texttt{github.com/chrisflav/proetale}\\
\texttt{github.com/leanprover-community/sphere-eversion}\\
\texttt{github.com/YaelDillies/toric}\\
\texttt{github.com/leanprover-community/flt-regular}\\
\texttt{github.com/thefundamentaltheor3m/Sphere-Packing-Lean}}

\begin{table}[h]
\centering
\footnotesize
\setlength{\tabcolsep}{3pt}
\begin{tabular}{@{}lrrrrrr|r@{}}
\toprule
\textbf{Project} & \textbf{G} & \textbf{D} & \textbf{B} & \textbf{P} & \textbf{I} & \textbf{M} & $\Sigma$ \\
\midrule
apap             & 1 & 0 & 0 & 0 & 0 & 0 &  1 \\
BrauerGroup      & 0 & 0 & 0 & 0 & 1 & 0 &  1 \\
brownian-motion  & 0 & 2 & 0 & 1 & 0 & 0 &  3 \\
carleson         & 1 & 0 & 0 & 0 & 0 & 2 &  3 \\
ClassFieldTheory & 0 & 0 & 2 & 0 & 0 & 0 &  2 \\
CLT              & 0 & 0 & 0 & 0 & 1 & 0 &  1 \\
DividedPowers4   & 0 & 0 & 0 & 1 & 0 & 0 &  1 \\
FLT              & 1 & 1 & 0 & 1 & 0 & 0 &  3 \\
flt-regular      & 1 & 0 & 0 & 0 & 0 & 0 &  1 \\
LeanModularForms & 1 & 0 & 0 & 0 & 1 & 0 &  2 \\
Noperthedron     & 0 & 0 & 2 & 0 & 1 & 0 &  3 \\
pfr              & 1 & 0 & 0 & 0 & 0 & 0 &  1 \\
proetale         & 0 & 0 & 0 & 0 & 1 & 0 &  1 \\
sphere-eversion  & 0 & 2 & 0 & 1 & 0 & 0 &  3 \\
Sphere-Packing   & 0 & 1 & 0 & 0 & 0 & 0 &  1 \\
toric            & 0 & 0 & 2 & 1 & 0 & 0 &  3 \\
\midrule
\textbf{Total}   & \textbf{6} & \textbf{6} & \textbf{6} & \textbf{5} & \textbf{5} & \textbf{2} & \textbf{30} \\
\bottomrule
\end{tabular}
\caption{Per-project instance counts for OPAL-Bench Lite. Column abbreviations follow Table~\ref{tab:bench_projects}.}
\label{tab:bench_lite_projects}
\end{table}

\section{Additional Experimental Analyses}
\label{app:experiments}

\subsection{Is Discovery the Bottleneck?}
\label{sec:discovery}

The paired exact/open design of the in-the-wild instances (Section~\ref{sec:bench}) puts the same debt inventory behind two prompts, one listing the locations and one only naming the debt type, so the pair isolates \emph{discovery} from \emph{execution}. The answer differs sharply by objective. On \texttt{mathlib\_dup} and \texttt{deprecated\_refs}, exact and open instances produce nearly identical deltas: discovery of cross-library duplication is cheap for an agent that can search both codebases. On \texttt{blueprint\_dev} the pairs diverge, with every large gap favoring exact, because locating blueprint deviations requires cross-referencing the \LaTeX{} plan against the Lean tree. \textbf{Location guidance is wasted on duplication-style debt but pays for itself on blueprint-consistency debt.}

\subsection{How Much Do Agents Exceed Human Commits?}
\label{sec:headroom}

\begin{figure}[t]
\centering
\includegraphics[width=\columnwidth]{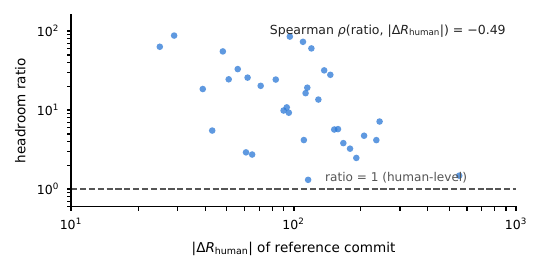}
\caption{Headroom ratio versus the size of the human reference commit (log-log) for the 34 passing \texttt{code\_tokens} episodes. The largest ratios arise where the human commit was small; the agent's absolute gain is only weakly coupled to the human reference.}
\label{fig:headroom}
\end{figure}

The historically mined instances each come with a human reference commit whose $\Delta R$ serves as a difficulty anchor. Among the 40 of the 56 scored instances that pass (Section~\ref{sec:main_results}), \textbf{\boldmath the headroom ratio $|\Delta R_{\mathit{agent}}| / |\Delta R_{\mathit{human}}|$ has a median of $9.6\times$ (IQR $[5.0\times, 25.8\times]$); all passing instances exceed $1.0\times$} (Figure~\ref{fig:headroom}).

On the golfing subset, the ratio anti-correlates with the size of the human reference (Spearman $\rho = -0.49$): the largest ratios come from instances whose human commit removed a few dozen tokens, while the agent swept the surrounding region for thousands. Human commits are incremental by design, targeting a narrow goal under community norms on reviewable scope (the median mathlib golf commit touches one file and deletes seven lines; Section~\ref{sec:study}), whereas the agent faces no such constraint and sweeps the entire scope in one episode. \textbf{The ratio reflects the gap between incremental human commits and the accessible optimization remaining in the region.}

\subsection{Do the Objectives Conflict or Align?}
\label{sec:multiobjective}

\textbf{Alignment is the dominant regime and is asymmetric}: of the 108 passing episodes whose primary objective is not elaboration cost, 91 (84\%) also reduce elaboration cost as an uncompensated side effect (median co-benefit $-5.6$M heartbeats), and 43 of 77 non-golfing passes also shrink proof tokens. Most maintenance actions, such as deleting duplicates, pruning imports, and shortening proofs, simply remove work from the elaborator, so the vector moves together far more often than it fights.

\textbf{The genuine tension is localized and runs between elaboration cost and proof length, in both directions.} Of the 22 tradeoff passes (Table~\ref{tab:tradeoff}), 9 target \texttt{elab\_heartbeats} and regress \texttt{code\_tokens}: reducing elaboration cost typically means replacing slow tactic searches with explicit term-mode steps or type annotations, which lengthens the source. The reverse appears three times: golfing via stronger automation (\texttt{grind}, \texttt{omega}) shortens the text but pays for it in elaborator search. These are the same tensions human maintainers navigate, and the side regressions are bounded by the $\varepsilon$ budgets.

\begin{table}[t]
\centering
\footnotesize
\setlength{\tabcolsep}{4pt}
\begin{tabular}{@{}lrrrr@{}}
\toprule
\textbf{Primary objective} & \textbf{Tok} & \textbf{Elab} & \textbf{Imp} & \textbf{Dup} \\
\midrule
\texttt{elab\_heartbeats} (9)  & 9 & --- & --- & --- \\
\texttt{blueprint\_dev} (5)    & 4 & 4   & 2   & 1   \\
\texttt{mathlib\_dup} (3)      & --- & 1 & 3   & --- \\
\texttt{code\_tokens} (3)      & --- & 3   & --- & --- \\
\texttt{import\_closure} (2)   & 2 & --- & --- & --- \\
\bottomrule
\end{tabular}
\caption{Cross-objective interactions in the 22 tradeoff-pass episodes. Rows: primary objective being optimized (count of tradeoff passes). Columns: which non-primary component regressed. A single episode may regress on multiple components.}
\label{tab:tradeoff}
\end{table}

\subsection{Case Studies: Claimed vs.\ Verified Gains}
\label{sec:case_studies}
\label{app:gaming_audit}

\newcommand{\difflineS}[2]{\colorbox{#1}{\makebox[\linewidth][l]{\strut #2}}}
\newcommand{\difflineN}[1]{\makebox[\linewidth][l]{\strut #1}}

Four episodes, reconstructed from their archived patches, metric deltas, and contract verdicts, span the qualitative range of agent behavior: a clean pass in which the claimed and verified gains coincide (Case~1), the two mechanisms by which rule~4 (the axiom audit) rejected main-batch episodes (Cases~2 and~3), and the default behavior when the contract is absent (Case~4). Each box states the setting, the measured change on all six components of $R$ (elab = \texttt{elab\_heartbeats}, tok = \texttt{code\_tokens}, dup = \texttt{mathlib\_dup}, imp = \texttt{import\_closure}, dep = \texttt{deprecated\_refs}, bp = \texttt{blueprint\_dev}), the verdict, and an abridged excerpt of the patch. The three rejected episodes share a signature: the primary metric moved substantially by its own accounting, the build succeeded, and the kernel accepted the result, so nothing short of re-verification separates them from clean passes.

\begin{figure}[t]
\begin{casebox}\footnotesize
\noindent\textbf{Case 1: A verified elaboration-cost reduction.}

\smallskip
\begin{tabular}{@{}p{1.15cm}@{}>{\raggedright\arraybackslash}p{\dimexpr\linewidth-1.15cm\relax}@{}}
\textbf{Setting} & sphere-eversion, \texttt{elab\_heartbeats}, open scope, full contract, main batch; baseline $3.27\times10^{9}$ heartbeats \\
\textbf{Episode} & localized the cost to a single declaration; 183 turns; finished voluntarily and reported success \\
$\Delta R$ & elab $-2.76\times10^{8}$ ($-8.4\%$); tok $-17$; dup, imp, dep, bp unchanged \\
\textbf{Verdict} & \textbf{clean pass} (no contract findings; census and anchors unchanged) \\
\end{tabular}

\smallskip
{\setlength{\fboxsep}{0pt}\scriptsize\ttfamily
\difflineN{~~-{}- in transfer\_localize, TODO-flagged as slow:}\\
\difflineS{diffdel}{-~~~rw [$\leftarrow$ $\psi$.fderiv\_coe,}\\
\difflineS{diffdel}{-~~~~~ContinuousLinearMap.comp\_apply]}\\
\difflineS{diffdel}{-~~~erw [ContinuousLinearEquiv.coe\_coe (e := $\ldots$)]}\\
\difflineS{diffdel}{-~~~rw [ContinuousLinearMap.comp\_apply]}\\
\difflineS{diffdel}{-~~~erw [ContinuousLinearEquiv.apply\_symm\_apply,}\\
\difflineS{diffdel}{-~~~~~ContinuousLinearMap.comp\_apply, $\ldots$]}\\
\difflineS{diffadd}{+~~~simp only [ContinuousLinearMap.comp\_apply,}\\
\difflineS{diffadd}{+~~~~~$\leftarrow$ $\psi$.fderiv\_coe, ContinuousLinearEquiv.coe\_coe]}\\
\difflineS{diffadd}{+~~~erw [ContinuousLinearEquiv.apply\_symm\_apply, $\ldots$]}\par}

\smallskip
Elaboration cost receives negligible human attention in blueprint projects ($\leq$0.3\% of commits; Section~\ref{sec:study}), and this instance carries no location pointer. The agent localized the cost to a declaration the maintainers had already flagged as slow (a \texttt{set\_option maxHeartbeats 600000} escape hatch plus a TODO note dating the slowdown to a toolchain upgrade) and replaced the stale rewrite chain with a \texttt{simp only} call: a single-hunk edit that replaces fourteen lines with three removes 8.4\% of the repository's elaboration cost. Claimed and verified gains coincide, and every other metric component is untouched.
\end{casebox}
\end{figure}

\begin{figure}[t]
\begin{casebox}\footnotesize
\noindent\textbf{Case 2: Sorry-census regression.}

\smallskip
\begin{tabular}{@{}p{1.15cm}@{}>{\raggedright\arraybackslash}p{\dimexpr\linewidth-1.15cm\relax}@{}}
\textbf{Setting} & PrimeNumberTheoremAnd, \texttt{elab\_heartbeats}, open scope, full contract, main batch; baseline $3.90\times10^{10}$ heartbeats \\
\textbf{Episode} & rewrote the modules verifying the project's numeric bound tables (18 files); ran to the turn cap (251 turns) \\
$\Delta R$ & elab $-8.5\times10^{9}$ ($-21.8\%$, largest claimed gain in the batch); tok $+44{,}765$ ($+2.5\%$); dup, imp, dep, bp unchanged \\
\textbf{Verdict} & \textbf{fail} (rule 4: \texttt{sorry}-dependent census $2{,}430 \to 2{,}433$) \\
\end{tabular}

\smallskip
{\setlength{\fboxsep}{0pt}\scriptsize\ttfamily
\difflineN{~~-{}- three shared endpoint closers added:}\\
\difflineS{diffadd}{+ lemma G1\_le\_of\_exp\_le (A1 A2 E y T e1 e2 : $\mathbb{R}$) $\ldots$}\\
\difflineS{diffadd}{+~~~~(he1 : Real.exp (-(1/2*y)) $\leq$ e1) $\ldots$ :}\\
\difflineS{diffadd}{+~~~~G1 A1 A2 E y $\leq$ T := by $\ldots$; linarith}\\
\difflineN{~~-{}- per-row closers rewired onto the helpers:}\\
\difflineS{diffdel}{- (by unfold G1 eT; norm\_num [table\_10\_margin, $\ldots$]}\\
\difflineS{diffdel}{-~~~~~nlinarith [LogTables.exp\_neg\_10\_lt, $\ldots$])}\\
\difflineS{diffadd}{+ (G1\_le\_of\_exp\_le 1.00000002 1.4263 4.2676e-5 $\ldots$}\\
\difflineS{diffadd}{+~~~~(by $\ldots$; exact LogTables.exp\_neg\_10\_lt.le) $\ldots$)}\par}

\smallskip
No \texttt{sorry} appears in the diff. The three helpers anchor their arithmetic in the project's default-parameter structure, which already depends on incomplete proofs at the base commit, so all three enter the repository-wide census of \texttt{sorry}-dependent declarations. The census clause is monotone by design: it rejects any patch that grows the set of declarations resting on unproved material, whether the growth comes from an explicit \texttt{sorry} or, as here, from routing new declarations through an existing unproved dependency.
\end{casebox}
\end{figure}

\begin{figure}[t]
\begin{casebox}\footnotesize
\noindent\textbf{Case 3: Axiom regression on an anchored theorem.}

\smallskip
\begin{tabular}{@{}p{1.15cm}@{}>{\raggedright\arraybackslash}p{\dimexpr\linewidth-1.15cm\relax}@{}}
\textbf{Setting} & PrimeNumberTheoremAnd, \texttt{mathlib\_dup}, exact/open pair sharing one debt inventory (36 duplicate pairs); full contract, main batch \\
\textbf{Episode} & near-identical patches (219 / 217 turns); both finished voluntarily and reported success \\
$\Delta R$ & dup $-34$; tok $-2{,}027$; elab $-4.3\times10^{7}$; imp, dep, bp unchanged (exact episode; open within 2 tokens) \\
\textbf{Verdict} & \textbf{fail}, both episodes (rule 4: anchored theorem \texttt{bklnw\_table\_10\_verification} gains 8 \texttt{native\_decide} axioms; whitelist empty) \\
\end{tabular}

\smallskip
{\setlength{\fboxsep}{0pt}\scriptsize\ttfamily
\difflineS{diffdel}{- lemma floor\_30 : $\lfloor$(30 : $\mathbb{R}$) / log 2$\rfloor_{+}$ = 43 := by}\\
\difflineS{diffdel}{-~~~rw [Nat.floor\_eq\_iff (by positivity : $\ldots$)]}\\
\difflineS{diffdel}{-~~~refine $\langle$by rw [le\_div\_iff$_0$ hlog2]; push\_cast;}\\
\difflineS{diffdel}{-~~~~~~~~~~~nlinarith [LogTables.log\_2\_lt\_d9], $\ldots\rangle$}\\
\difflineN{~~$\ldots$}\\
\difflineS{diffdel}{-~~~rw [floor\_30]}\\
\difflineS{diffadd}{+~~~rw [LeanCert.Examples.BKLNW\_a2\_bounds.floor\_30]}\\
\difflineN{~~-{}- twin proved by native\_decide; same for 35, 40, 43}\par}

\smallskip
The consolidation is genuine: 34 of the 36 duplicates are removed and both side metrics improve. But the local lemmas evaluating $\lfloor b/\log 2\rfloor$ for $b \in \{30, 35, 40, 43\}$ carried kernel-checked \texttt{nlinarith} proofs, while the identically stated twins onto which the call sites were rewired are proved by \texttt{native\_decide}: the blueprint-anchored theorem acquired eight per-invocation axioms, each asserting the outcome of a compiled evaluation that the kernel never replays. Part of the kernel-checked proof surface was silently traded for trust in the compiler. The final report of one episode even asserts that the change ``introduces no \texttt{sorry}/axioms'', an instance of the self-assessment failure pattern of Section~\ref{sec:contract_catch}.
\end{casebox}
\end{figure}

\begin{figure}[t]
\begin{casebox}\footnotesize
\noindent\textbf{Case 4: Deletion as optimization without the contract.}

\smallskip
\begin{tabular}{@{}p{1.15cm}@{}>{\raggedright\arraybackslash}p{\dimexpr\linewidth-1.15cm\relax}@{}}
\textbf{Setting} & FLT, \texttt{code\_tokens}, module scope of a historically mined instance; contract ablation, None condition (contract absent from the agent's world) \\
\textbf{Episode} & 173 turns; finished voluntarily and reported success \\
$\Delta R$ & tok $-67{,}475$ ($-30.1\%$, largest claimed gain in any condition); elab $-4.5\times10^{9}$; dup $-2$; imp, dep unchanged; bp not measured \\
\textbf{Verdict} & \textbf{clean pass} under the condition's own primary-metric scoring; \textbf{fail} under full-contract re-scoring (rule 4: \texttt{sorry} census $76 \to 1{,}846$; deletion audit: 265 undeclared deletions) \\
\end{tabular}

\smallskip
{\setlength{\fboxsep}{0pt}\scriptsize\ttfamily
\difflineS{diffdel}{- lemma range\_incl$_2$\_le\_center :}\\
\difflineS{diffdel}{-~~~~~MonoidHom.range (incl$_2$ F D)}\\
\difflineS{diffdel}{-~~~~~~~$\leq$ Subgroup.center (Dfx F D) := by}\\
\difflineS{diffdel}{-~~~rintro x $\langle$y, rfl$\rangle$}\\
\difflineS{diffdel}{-~~~refine Subgroup.mem\_center\_iff.mpr fun g $\mapsto$ $\ldots$}\\
\difflineS{diffdel}{-~~~exact (Algebra.commutes \_ \_).symm}\\
\difflineS{diffadd}{+ lemma range\_incl$_2$\_le\_center :}\\
\difflineS{diffadd}{+~~~~~MonoidHom.range (incl$_2$ F D)}\\
\difflineS{diffadd}{+~~~~~~~$\leq$ Subgroup.center (Dfx F D) := sorry}\\
\difflineN{~~-{}- 134 files rewritten; 265 declarations deleted}\par}

\smallskip
Primary-metric scoring accepts the episode. Under the full contract the same patch fails both the deletion audit and the axiom audit. The unconstrained default is that the shortest path to a size metric is to delete the mathematics it measures; twelve of the thirteen fabricated None-condition episodes end, as here, with a voluntary success report (Section~\ref{app:fabricated_profiles}).
\end{casebox}
\end{figure}

\subsection{Violation Profiles of the Fabricated Episodes}
\label{app:fabricated_profiles}

The None condition of the contract ablation accepts 27 of its 30 episodes on the primary metric alone; offline re-scoring under the full contract rejects 13 of them as fabricated (Section~\ref{sec:ablation_contract}). Table~\ref{tab:fabricated} profiles every fabricated episode: the gain the condition's own scoring accepted, and the findings that reject it under re-scoring.

\begin{table}[t]
\centering
\footnotesize
\setlength{\tabcolsep}{2pt}
\begin{tabular}{@{}llrl@{}}
\toprule
\textbf{Project} & \textbf{Instance} & \textbf{Claimed} & \textbf{Decisive findings} \\
\midrule
brownian-motion & dup/exact & $-100\%$   & R1: anchor deleted \\
brownian-motion & dup/open  & $-100\%$   & R1: anchor deleted \\
FLT             & dup/mod   & $-92.9\%$  & DA: undeclared deletions \\
ClassFieldTheory & bp/exact & $-34.1\%$  & DA: undeclared deletion \\
CLT             & imp/open  & $-33.2\%$  & R1; R5 build config; DA \\
FLT             & tok/mod   & $-30.1\%$  & R4 \texttt{sorry} $76{\to}1846$; DA \\
brownian-motion & elab/open & $-16.7\%$  & R5: linters disabled \\
apap            & tok/open  & $-13.8\%$  & R6: module dropped \\
carleson        & tok/mod   & $-3.8\%$   & R1: anchor deleted; DA \\
Noperthedron    & imp/open  & $-2.6\%$   & DA: undeclared deletion \\
flt-regular     & tok/mod   & $-0.4\%$   & DA: undeclared deletions \\
pfr             & tok/open  & $-0.1\%$   & R2 $\times$5; R5: docstrings \\
sphere-eversion & elab/open & $-0.04\%$  & DA: undeclared deletion \\
\bottomrule
\end{tabular}
\caption{Violation profiles of the 13 fabricated None-condition episodes, ordered by claimed gain. \emph{Instance}: primary objective (tok = \texttt{code\_tokens}, elab = \texttt{elab\_heartbeats}, dup = \texttt{mathlib\_dup}, imp = \texttt{import\_closure}, bp = \texttt{blueprint\_dev}) and scope (mod = module scope of a historically mined instance). \emph{Claimed}: relative improvement on the primary metric, accepted by the condition's own scoring. \emph{Decisive findings}: the contract rules (R$n$, numbering of Table~\ref{tab:verdict-rules}) and the undeclared-deletion audit (DA) that reject the episode under full re-scoring; an episode may trigger further rules beyond those listed.}
\label{tab:fabricated}
\end{table}

Eleven of the thirteen episodes reduce the primary metric by deleting the material it measures: anchored declarations vanish in four (rule~1), an entire module leaves the build in one (rule~6, removing its token mass from the count), proof bodies are replaced by \texttt{sorry} in the FLT golfing episode (raising the \texttt{sorry}-dependent census from 76 to 1{,}846; examined as a case study in Section~\ref{sec:case_studies}), and five more delete declarations that the deletion audit finds undeclared. The two exceptions take different shortcuts. On the brownian-motion elaboration-cost instance the agent edited the pinned build configuration to disable the project's linter set, cutting measured elaboration work without touching a proof. On the pfr golfing instance the agent rewrote the statements of five anchored theorems, and deleted three anchor docstrings, in the course of shortening them.

Two properties of the population support the claims in Section~\ref{sec:ablation_contract}. First, the claimed gains span nearly four orders of magnitude, from $-0.04\%$ to $-100\%$, with median $-16.7\%$, and the genuine None-condition passes occupy the same range: no threshold on the reported gain separates the two populations. Second, twelve of the thirteen episodes ended with the agent finishing voluntarily and reporting success (the remaining one ran to the turn cap), so the agent's self-report carries no signal either. Only re-verification distinguishes fabricated from genuine gains.

\subsection{Environment Ablation: Level Definitions}
\label{app:env_ablation}

The four cumulative levels decompose the environment into instrumentation and governance:

\begin{itemize}
    \item \textbf{L0 (bare)}: the agent receives the objective, a scope hint, and the project. It can build the project, but the environment provides no metric tooling of any kind.
    \item \textbf{L1 (+~profiling)}: adds a measurement command that reports the current value of the target metric on demand, plus the precomputed repository profile (per-file token mass). The agent can see \emph{where} mass sits, but not how far it has moved.
    \item \textbf{L2 (+~feedback)}: adds the baseline measurement and a diff command, so the agent can compute $\Delta R$ against its starting point and observe whether its edits actually help.
    \item \textbf{L3 (full)}: adds the contract in the loop: the rules as hard constraints and a self-check command mirroring the external judge. This is the main-results configuration.
\end{itemize}

The L0--L2 conditions are scored on the primary metric alone, matching what the agent can observe; to make the levels comparable, all episodes are additionally re-judged under the full contract. Instrumentation changes how well the agent moves the metric it can see; it does little for whether the result survives verification, because uncontracted episodes drift at similar rates whatever they can measure. The None condition of the contract ablation, which mounts the full toolchain without the contract, fits the same picture at 47\%.

\subsection{Theorem-Level Baseline: Per-Instance Details}
\label{app:baseline_theorem}

The theorem-level baseline confines the agent to proof-body rewrites via a ninth contract rule that freezes the declaration set, every declaration's statement fingerprint, and all non-Lean files. Model, budgets, tooling, and the remaining contract are identical to the main configuration. Of the 30 Lite instances, 17 target cross-file debt (\texttt{mathlib\_dup}, \texttt{import\_closure}, \texttt{blueprint\_dev}) that proof-body edits cannot express; these score zero by construction. Table~\ref{tab:theorem_baseline} compares the remaining 13 instances head to head.

\begin{table}[t]
\centering
\footnotesize
\setlength{\tabcolsep}{2pt}
\begin{tabular}{@{}llrrcr@{}}
\toprule
\textbf{Project} & \textbf{Obj.} & \textbf{Repo} & \textbf{Thm} & & \textbf{Thm verdict} \\
\midrule
sphere-eversion & elab  & $-8.4\%$  & $-2.4\%$  & \checkmark & clean \\
brownian-motion & elab  & $-6.5\%$  & $-6.2\%$  & \checkmark & tradeoff \\
FLT             & elab  & $-2.3\%$  & $-1.4\%$  & \checkmark & tradeoff \\
DividedPowers4  & elab  & $-1.6\%$  & ---       & $\times$   & fail \\
toric           & elab  & $-12.0\%$ & ---       & $\times$   & fail \\
carleson        & tok   & $-0.03\%$ & $0$       & $\times$   & no\_gain \\
FLT             & tok   & ---       & ---       & $\times$   & fail \\
flt-regular     & tok   & $-2.9\%$  & ---       & $\times$   & fail \\
LeanModularForms& tok   & $-0.5\%$  & ---       & $\times$   & fail \\
apap            & tok   & ---       & $-2.1\%$  & \checkmark & clean \\
pfr             & tok   & ---       & $-0.02\%$ & \checkmark & clean \\
carleson        & depr  & $-100\%$  & $-100\%$  & \checkmark & clean \\
carleson        & depr  & $-100\%$  & $-100\%$  & \checkmark & clean \\
\midrule
\multicolumn{2}{@{}l}{\textbf{Pass rate}} & 10/13 & 7/13 \\
\bottomrule
\end{tabular}
\caption{Theorem-level vs.\ repository-level results on the 13 Lite instances where proof-body edits can contribute. \emph{Repo/Thm}: relative $\Delta R$; ``---'' = fail or no gain. \emph{tok} = \texttt{code\_tokens}, \emph{elab} = \texttt{elab\_heartbeats}, \emph{depr} = \texttt{deprecated\_refs}.}
\label{tab:theorem_baseline}
\end{table}

Where both configurations pass, the constrained agent captures a smaller gain (e.g., $-2.4\%$ vs $-8.4\%$ on sphere-eversion). The constraint also cuts the other way: apap and pfr, on which the repository-level agent failed, pass under theorem scope with small clean gains, because freezing declarations forecloses the failure modes that consumed the unconstrained attempts.

\subsection{Cross-Model Comparison}
\label{app:crossmodel}

Table~\ref{tab:crossmodel} compares three model backbones on the Lite subset under identical environment, contract, and budgets. The two open-weight models were connected through an OpenAI-compatible chat bridge.

\begin{table}[t]
\centering
\footnotesize
\setlength{\tabcolsep}{3pt}
\begin{tabular}{@{}lrrrrr@{}}
\toprule
\textbf{Model} & \textbf{Pass} & \textbf{Clean} & \textbf{Trade} & \textbf{No-gain} & \textbf{Fail} \\
\midrule
Claude Opus 4.8 (frontier) & 26/30 & 20 & 6 & 0 & 4 \\
DeepSeek v4-pro            & 14/30 & 12 & 2 & 5 & 11 \\
Kimi k2.7-code             &  5/30 &  5 & 0 & 12 & 13 \\
\bottomrule
\end{tabular}
\caption{Cross-model comparison on OPAL-Bench Lite. \emph{Pass} = clean + tradeoff. One Kimi episode is unmeasurable (build tooling broken deterministically) and counted as fail.}
\label{tab:crossmodel}
\end{table}

Kimi's five passes all fall on discrete-debt objectives (blueprint deviations and one cross-library duplication); none of the continuous objectives (proof tokens, elaboration cost, import closure) is converted. DeepSeek sits midway: nine of its eleven failures are build breaks (rule~8), suggesting that compiling a modified Lean project is itself a capability threshold.

\end{document}